\documentclass{article}
\usepackage[preprint]{neurips_2024}

\usepackage[utf8]{inputenc}
\usepackage[T1]{fontenc}
\usepackage{hyperref}
\usepackage{url}
\usepackage{booktabs}
\usepackage{amsfonts}
\usepackage{nicefrac}
\usepackage{microtype}
\usepackage{xcolor}
\usepackage{graphicx}
\setcitestyle{square}
\usepackage{svg}
\usepackage{wrapfig}
\setcitestyle{numbers}

\title{Extracting Paragraphs from LLM Token Activations}

\author{%
  Nicholas Pochinkov \\
  Independent \\
  \texttt{work@nicky.pro} \\
  \And
  Angelo Benoit\\
  Independent\\
  \texttt{angelogbenoit@gmail.com} \\
  \And
  Lovkush Agarwal\\
  Independent\\
  \texttt{lovkush@gmail.com} \\
  \AND
  Zainab Ali Majid\\
  Independent\\
\texttt{zainab\_majid@hotmail.com} \\
  \And
  Lucile Ter-Minassian \\
  University of Oxford \\
  \texttt{lucile.ter-minassian@spc.ox.ac.uk}
}

\begin{document}
\maketitle
\begin{abstract}
Generative large language models (LLMs) excel in natural language processing tasks, yet their inner workings remain underexplored beyond token-level predictions. This study investigates the degree to which these models decide the content of a paragraph at its onset, shedding light on their contextual understanding. By examining the information encoded in single-token activations, specifically the "\textbackslash n\textbackslash n" double newline token, we demonstrate that patching these activations can transfer significant information about the context of the following paragraph, providing further insights into the model's capacity to plan ahead. %representation of context and structure.
\end{abstract}

\section{Introduction}

Recent advancements in large language models have revolutionized Natural Language Processing, enabling unprecedented performance on a wide range of tasks, including machine translation \cite{vaswani2017attention, raffel2020exploring}, question answering \cite{devlin2019bert,brown2020language}, and text generation \cite{radford2019language,brown2020language}. Despite these successes, our understanding of how these models internally process and represent information remains limited \citep{olah-feature-vis, circuits-ioi-interpretability-redundancy}.

Previous studies have demonstrated that internal model representations can reveal how models plan ahead in text generation. By \emph{intervening} on neural activations—specifically by patching them between different locations at inference time - we can uncover existing causal relationships \cite{zou2023representation, plan-future-tokens, act-add, subramani2022extracting, hernandez2023inspecting}. For instance, \textit{Pal et al.} use causal intervention methods in their \textit{Future Lens} approach \cite{future-lens} to show that individual hidden states at position $t$ contain signals rich enough to predict future tokens at $t + 2$ or beyond, and this insight has been used to improve performance 
of models \citep{gloeckle2024better,speculative-streaming}. However, existing interpretability research predominantly focuses on token-level predictions by examining how models predict individual words or tokens \citep{unlocking-the-future}, rather than exploring broader contexts such as the thematic coherence of a sentence or paragraph.

Our work aims to bridge the gap between token-level and paragraph-level understanding by investigating whether the information content of single-token activations remains relevant when we consider sequences of tokens, with a specific focus on the "\textbackslash n\textbackslash n" double newline token. We hypothesize that these activations contain information about the structure and content of the following paragraph, providing insight into the model's comprehension of larger textual units.

In section \ref{sec:observing}, we demonstrate through a preliminary experiment that text structure is embedded in a language model's attention scores. In section \ref{sec:experimenting}, we examine the extent to which a model, at the start of a paragraph, has already planned the rest of the generated text. To explore this, we patch activations onto a model with a neutral prompt -- a double newline -- and investigate whether the future paragraph contains information transferred at the hidden representation level. The code for our experiments is \href{https://anonymous.4open.science/r/extracting-paragraphs-65CF/README.md}{available anonymously}. Compute details can be found in Appendix \ref{sec:exp_details}. 

\newpage

\section{Is Text Structure Encoded in the Model's Attention Patterns?} \label{sec:observing}% Observing Language Model Attention Patterns

To motivate our approach, we first demonstrate that sequences of paragraphs can be identified through the analysis of an LLM's attention activations. We generate texts by prompting a model with instructions phrased as: "Tell me about topic 1 in $k$ words \verb|\n\n| tell me about topic 2 in $k$ words." These generated texts, referred to as \textit{original} contexts, are structured uniformly by instructing the model not to generate headings and additional comments. We then extract and inspect the combined attention patterns across all heads for each model-generated text. To observe the context switch, we conduct two key analyses, averaging across the textual generations: (1) the distribution of attention weights close to the topic change, and (2) the cosine similarity of attention output activations inside and across paragraphs, or topics. Experiment (1) checks to what extent attention heads focus on the current paragraph, whilst (2) investigates if attention outputs differ between paragraphs.
 % This analysis builds upon techniques such as the logit lens \citep{logit-lens}  and other interpretability methods \citep{olah-branch-specialization, DBLP/corr/abs-2108-13138}.

\begin{figure}[ht]
    \vspace{7pt}
    \centering
    \begin{minipage}{0.45\textwidth}
        \centering
        \includegraphics[width=\textwidth]{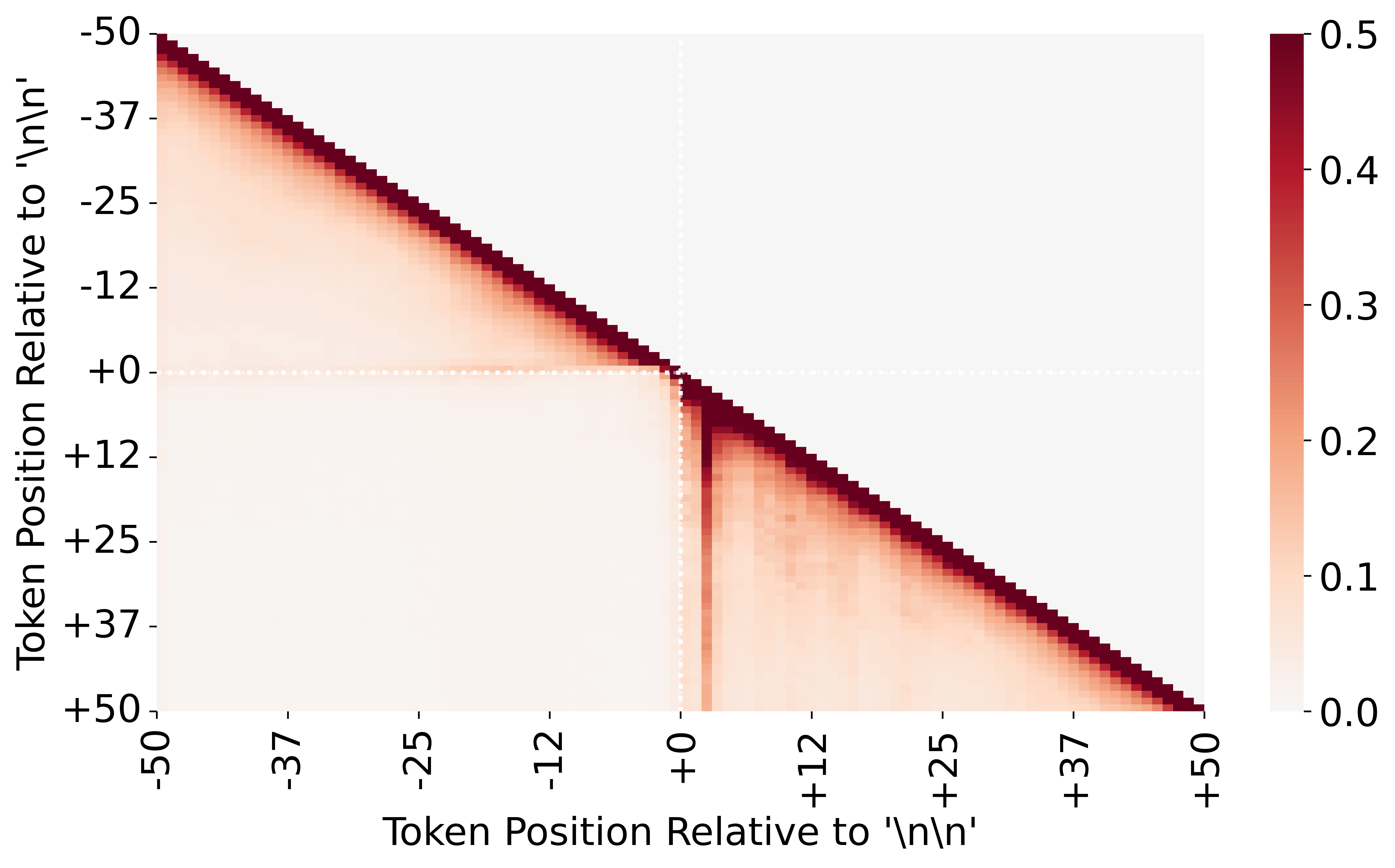}
    \end{minipage}\hfill
    \begin{minipage}{0.52\textwidth}
        \centering
        \includegraphics[width=\textwidth]{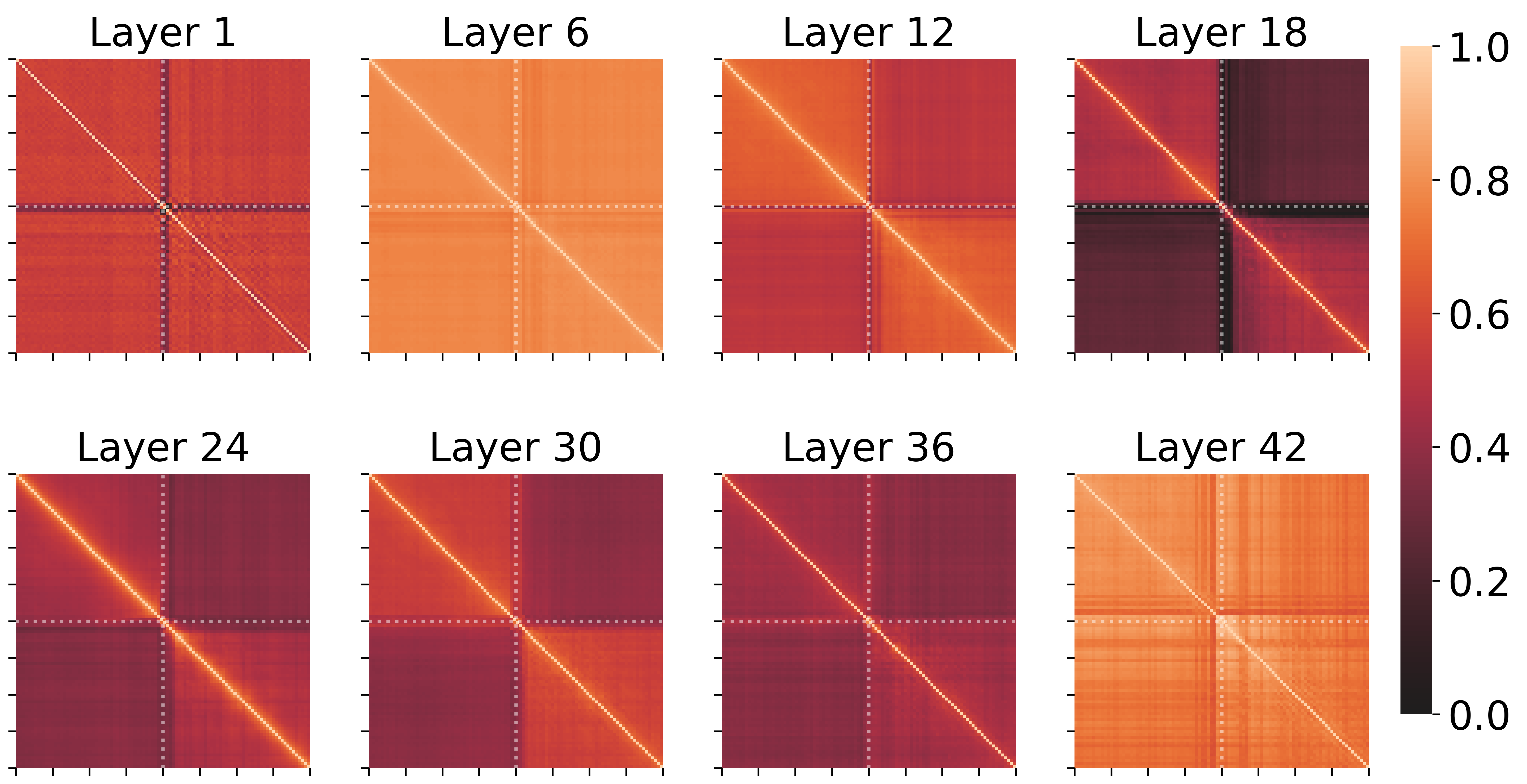}
    \end{minipage}
    \caption{(Left): Heat map of the average attention weights around the topic change. (Right): Cosine similarity between attention activations. Results averaged over 1,000 model-generated original contexts, sharing a common structure.}
    \label{fig:observation_exp}
\end{figure}
\vspace{7pt}

%The attention scores reveal a significant concentration of attention on the current paragraph across all model-generated texts.
Figure \ref{fig:observation_exp} shows the results of our attention pattern analysis. In our study, we used 20 prompts (i.e., pairs of topics), generating 50 texts per prompt, for a total of 1,000 generated texts. The generations were implemented with the Gemma 2, 9b model at a temperature of 0.3, and activations were retrieved using the HuggingFace Transformers library \cite{huggingface-transformers}. On the left, the attention weights indicate that the model tends to attend to previous tokens almost exclusively from the same paragraph. On the right, the cosine similarities of attention outputs show how strongly text structure is encoded across various layers. In the first 18 layers, the cosine similarities of attention activations increase within paragraphs and decrease across paragraphs, suggesting that the model is learning abstract representations in early layers, where it gradually develops an understanding of the paragraph topic. Another consistent finding across all experimental settings is that distinctions between paragraphs diminish in the final layers, from layer 30 onwards. We conjecture that this may be due to the model eventually producing text of a very similar overall form for both topics. An additional plot displaying the cosine similarities for all 42 model layers can be found in Appendix \ref{sec:transferred}. Altogether, our preliminary experiments suggest that our model maintains a strong contextual awareness during text generation, in line with research allowing consistent text embed fine-tuning \citep{llms-universal-embedders}, and "planning" \citep{predicting-vs-acting, plan-future-tokens, janus2023llmmyopia}. These results also confirm that the context switch at the start of a paragraph is encoded in the activation space.
%\nicky{(Idk how to write this compactly)}

\begin{figure}[ht]
\vspace{7pt}
\centering
\includegraphics[width=.8\textwidth]{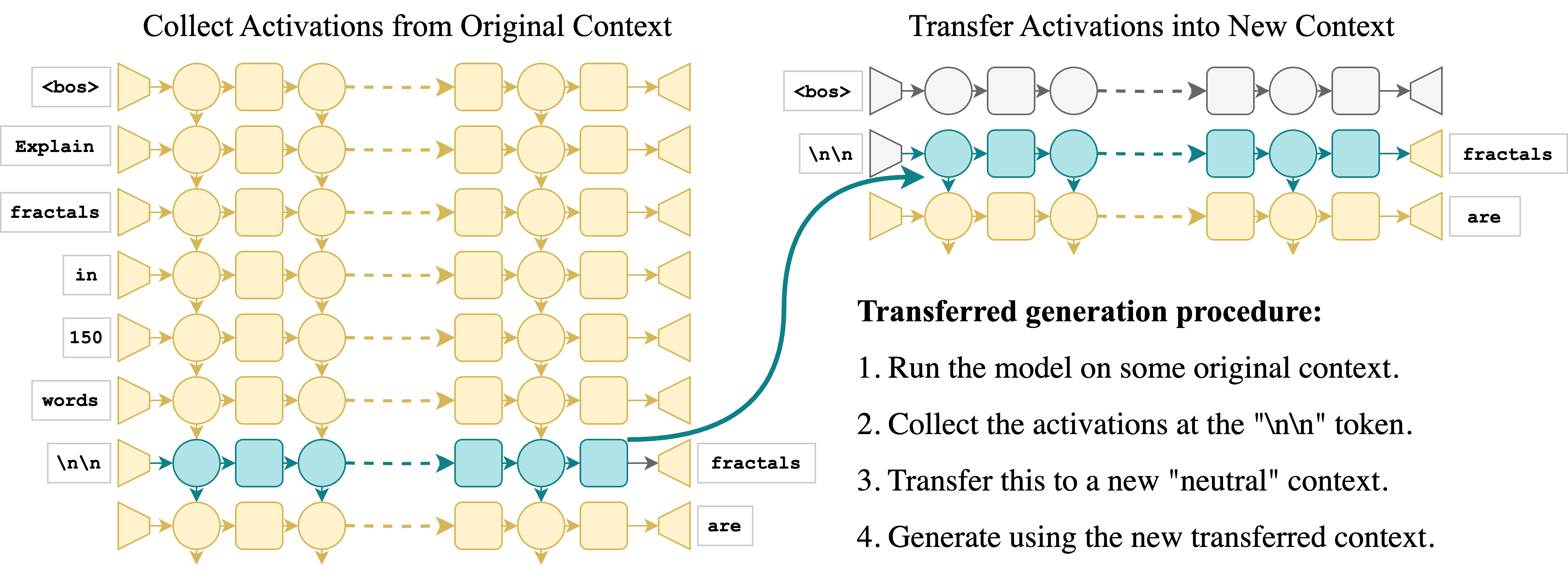}
\caption{Diagram describing our approach. After collecting activations at the transition token on the original context model, we transfer these to all layers of the neutrally-prompted model.}
\label{fig:infographic}
\end{figure}
\vspace{7pt}

\section{Generation Experiments with Transferred Activations} \label{sec:experimenting}

% To investigate how models plan ahead for a new section, we conduct a series of generation experiments, illustrated in Figure \ref{fig:infographic}. We first prompt a model with the \textit{original} contexts (i.e., pairs of topics) and extract the activations of the double newline token between topics. These activations are then transferred to the corresponding double newline token of a second, fresh model. This transfer occurs across all layers of the model, i.e.\ including both attention and Multi-Layer Perceptron (MLP) layers. We then prompt our transferred model with a neutral prompt, specifically the double newline token. Doing so effectively "seeds" the fresh model with information encoded solely in the \lucile{saying single token may be confusing here?} single-token activation vector of the double newline transition, without additional context. Our goal is to analyze how much of the second paragraph's information is contained in this token, assessing the extent to which the model has planned the rest of the generated text at the start of a new paragraph. 

\begin{wrapfigure}{r}{0.5\textwidth}
  \centering
  \includegraphics[width=.5\textwidth]{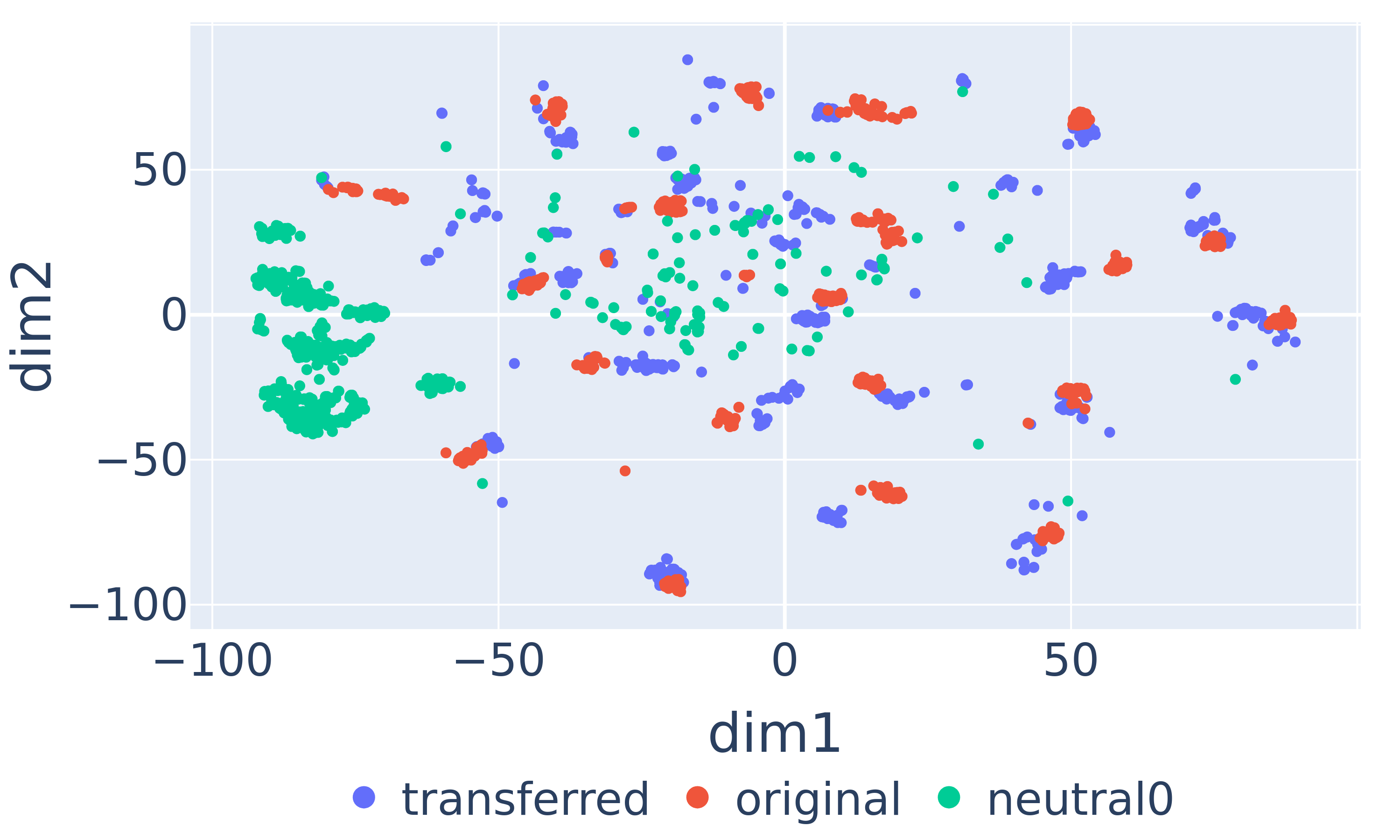}
  \caption{Context similarity visualised with T-SNE. Results over 1,000 original contexts.}
  \label{fig:tsne}
\end{wrapfigure}

To investigate how models plan ahead for a new section, we conduct a series of generation experiments, illustrated in Figure \ref{fig:infographic}. We first prompt a model with the \textit{original} contexts (i.e., pairs of topics) and extract the activations of the double newline token between topics. These activations are then transferred to the corresponding double newline token of a neutral context, i.e.\ a fresh model ``neutrally'' prompted with "<bos>\textbackslash n\textbackslash n". This transfer occurs across all layers of the model, i.e.\ including both attention and Multi-Layer Perceptron (MLP) layers. Doing so effectively "seeds" the neutral context with information encoded solely in the activation vector of the double newline token, without additional context. Our goal is to analyze how much of the second paragraph's information is contained in this token, assessing the extent to which the model has planned the rest of the generated text at the start of a new paragraph.

%\angelo{could be useful to have an established alias for newline-newline token} #lucile: double newline does it
To analyze the context of the transferred generations—i.e., texts generated from neutrally prompted models with transferred activations—we use state-of-the-art sentence embedding techniques \citep{beir-text-embedding-retrieval, mteb-text-embedding, uae-large-text-embed-model}. We convert the output sequence of tokens into a single activation vector using ALL MPNET Base v2 \citep{reimers-2019-sentence-bert, song2020mpnet}, and compare the semantic similarity between the \emph{original} generations and those produced from the \emph{transferred} activations. Additionally, we compare these with texts generated by the model using the same neutral prompt without activation transplantation, referred to as the \emph{neutral0} generations. (The relevance of `0' is explained two paragraphs below.)

\begin{wrapfigure}{r}{0.5\textwidth}
  \centering
  % Image
  \includegraphics[width=\linewidth]{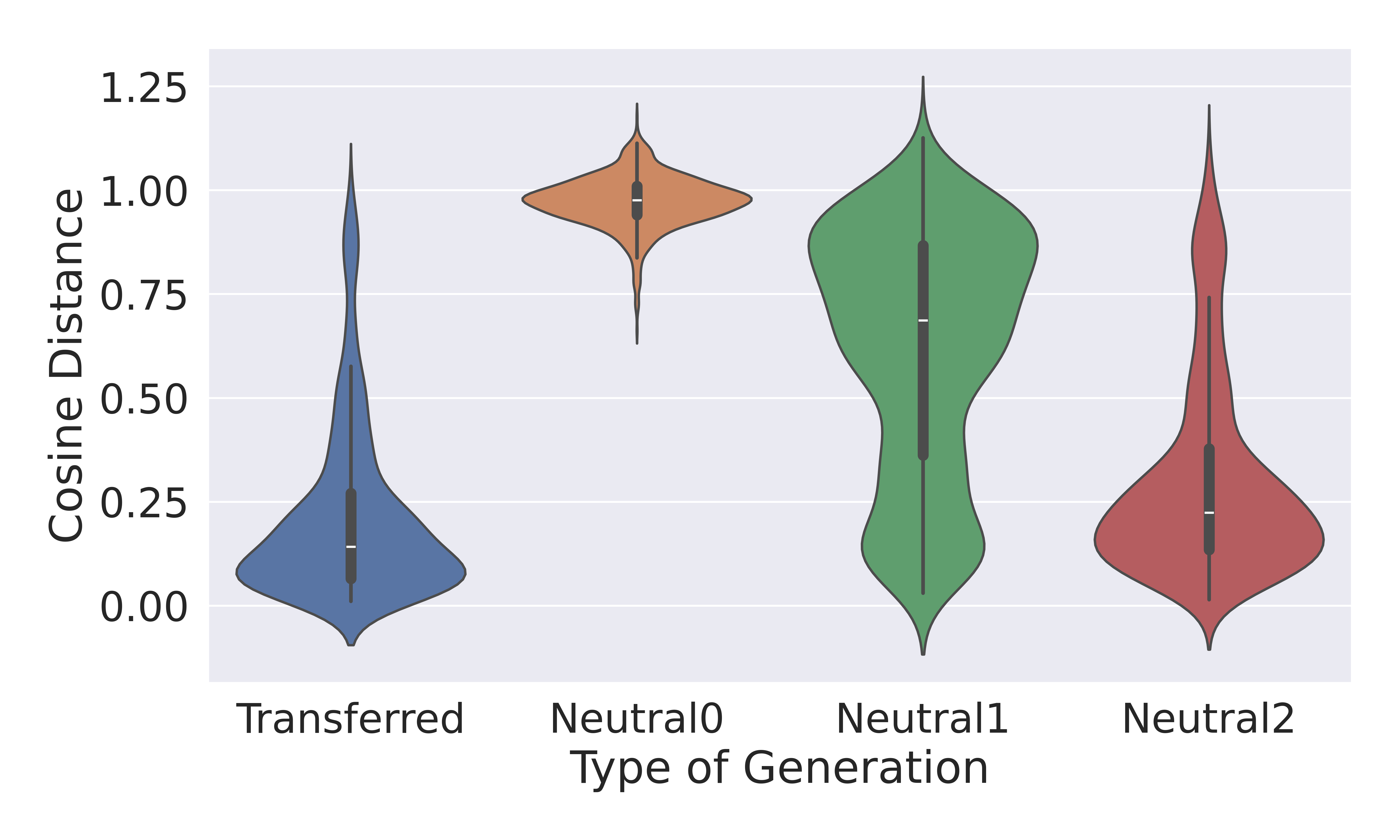}
  \vspace{-5pt} % Reduce space between image and table (negative value)
  
  % Table
  \begin{tabular}{lll}
  Type of Generation  & Neutral2 & Transferred   \\ \hline
  \begin{tabular}[c]{@{}l@{}}Mean \\ cosine distance \\ to original (Std)\end{tabular} & \begin{tabular}[c]{@{}l@{}} 0.303 \\ (0.239) \end{tabular}    & \begin{tabular}[c]{@{}l@{}} 0.214 \\ (0.210) \end{tabular}  \\ \hline
  \end{tabular}
  \caption{Distribution of cosine distances to the original generation. Contexts are summarized using sentence transformers, and distributions are taken over 1,000 original contexts.}
  \label{fig:violin}
\end{wrapfigure}

Figure \ref{fig:tsne} shows the first two dimensions of a T-Distributed Stochastic Neighbor Embedding (T-SNE) for the neutral0, original, and transferred generated texts. Our findings reveal a remarkable degree of semantic similarity between the original paragraphs and those generated from the transferred double newline token activations. For each prompt, the \emph{transferred} cluster aligns well with the \emph{original} cluster. In contrast, texts generated from the unaltered, neutrally prompted model are randomly scattered, showing low similarity with the original generations. This confirms that the activations of the double newline token hold a lot of information about the upcoming paragraph despite it being a separate topic from the previous one. An additional plot comparing the generations with PHATE can be found in the Appendix \ref{sec:transferred}.

Given that we transfer the activations from every layer, we in particular transfer the activations of the final layer. This means that we are effectively telling the model what the next token is, so comparing this case with the neutral prompt may unfairly advantage our transferred generations. To address this, we add ``cheat'' tokens to assist the neutral baseline by hinting at the context of the next paragraph and removing the next-token prediction advantage of the transferred generations. Specifically, we create two additional sets of generations—\textit{neutral1} and \textit{neutral2}—where the neutral prompt is concatenated with one or two ``cheat'' tokens, which are the first words of the following paragraph in the original generation. We also use the same sentence transformer to retrieve the embeddings of the generations.

Figure \ref{fig:violin} displays the distribution of cosine distances to the original generation, comparing the transferred generations with \textit{neutral0}, \textit{neutral1}, and \textit{neutral2}. The transferred generations significantly outperform \textit{neutral0} and \textit{neutral1}, being much closer to the original generations. Notably, \textit{neutral2} generations are much closer to the original than \textit{neutral0} and \textit{neutral1}, and share a similarly shaped distribution to the transferred generations. However, as shown in the table comparing the average cosine distances, the transferred generations still outperform \textit{neutral2} in terms of closeness to original - with a t-test of the two sets of cosine distances rendering $p < 0.001$. Exhaustive results can be found in Table \ref{tab:furtherres_avg} in the Appendix.

\section{Related work} 
Our work leverages activation patching, a standard technique in mechanistic interpretability introduced by \textit{Vig et al.} \cite{vig2020investigating, zhang2023towards, DBLP:conf/emnlp/GevaSBL21}. This is in line with the work of \textit{Jenner et al.}, which uses activation patching to understand the look-ahead behavior in Leela Chess Zero's policy network. However, contrasting with their focus on chess strategy optimization, we explore how language models anticipate the context of future paragraphs. Our work also builds upon recent studies examining lookahead in causal language models, although they adopt a different approach to the question. For example, the Future Lens approach \citep{future-lens} investigates how much signal individual hidden states contain by using them to predict subsequent embeddings, and \textit{Wu et al.} \citep{plan-future-tokens} modify training procedures to gain deeper insights into token planning. Addressing context planning from another perspective, the \textit{Patchscope} approach \cite{ghandeharioun2024patchscope} combines inspection prompts and activation engineering to investigate how context is \emph{read}, demonstrating that this process occurs predominantly in early layers. 

While these studies offer valuable perspectives, our research question diverges by examining the phenomenon at a different scale. We focus on the \emph{context} of an entire section of generated text, notably using sentence embedding. This approach allows us to explore broader contextual relationships and planning mechanisms within language models.
\citep{janus2023llmmyopia}
.

\section{Discussion}

\textbf{Conclusion}

In this work, we investigate how an LLM plans for future context. By using a specific set of prompts, we observed the model's attention allocation between the current paragraph and the previous paragraph on a different topic. We further examined the extent of ``pre-planned'' information the model holds for the subsequent paragraph by performing activation transfers on a neutrally prompted model.
Our findings suggest that a single token encodes a substantial amount of contextual information about the forthcoming section, and that most (but not all) of this information seems to be contained in the first two tokens of generation.

\textbf{Limitations} 
\label{sec:limitations}
While our framework provides valuable insights, it has several limitations. First, it is designed specifically for autoregressive (or “causal”) models and does not apply to word2word models, which lack the same sequential generation process. Thus, its utility is tied to autoregressive model architectures. Second, this study is an experimental investigation rather than a comprehensive solution. The methods are not foolproof and are not suited for explaining sensitive or high-stakes models. This work should be viewed as a foundational step for future research aimed at developing more robust methods. Additionally, our experiments have been conducted using only a specific language model. However, we reasonably expect that our findings generalize to other transformer-based language models, which are widely used today.

\textbf{Future work} Our experiments focus on abrupt context switches, which, while useful for certain analyses, may not fully represent realistic scenarios. Future work could explore applying our approach to cohesive texts without abrupt context switches, though this poses its own challenges. Specifically, distinguishing between the information the model ``remembers'' across sections and what it knows at the onset of a paragraph is complex. Currently, our framework examines the model's planning for the ``next paragraph'', but future research could extend this to predict further sections ahead. Evaluating whether the framework can anticipate not just the immediate next paragraph but also subsequent sections could provide insights into its ability to construct and maintain a global narrative structure.
 
%This would help us understand how models maintain and develop context over longer spans of text, revealing limitations of our framework in more realistic scenarios. 
\newpage

\bibliography{paper}
\bibliographystyle{plainnat}

%%%%%%%%%%%%%%%%%%%%%%%%%%%%%%%%%%%%%%%%%%%%%%%%%%%%%%%%%%%%
\newpage
\appendix
\section{Further Experimental Results}
\label{sec:transferred}

\begin{figure}[ht]
\centering
\includegraphics[width=0.99\textwidth]{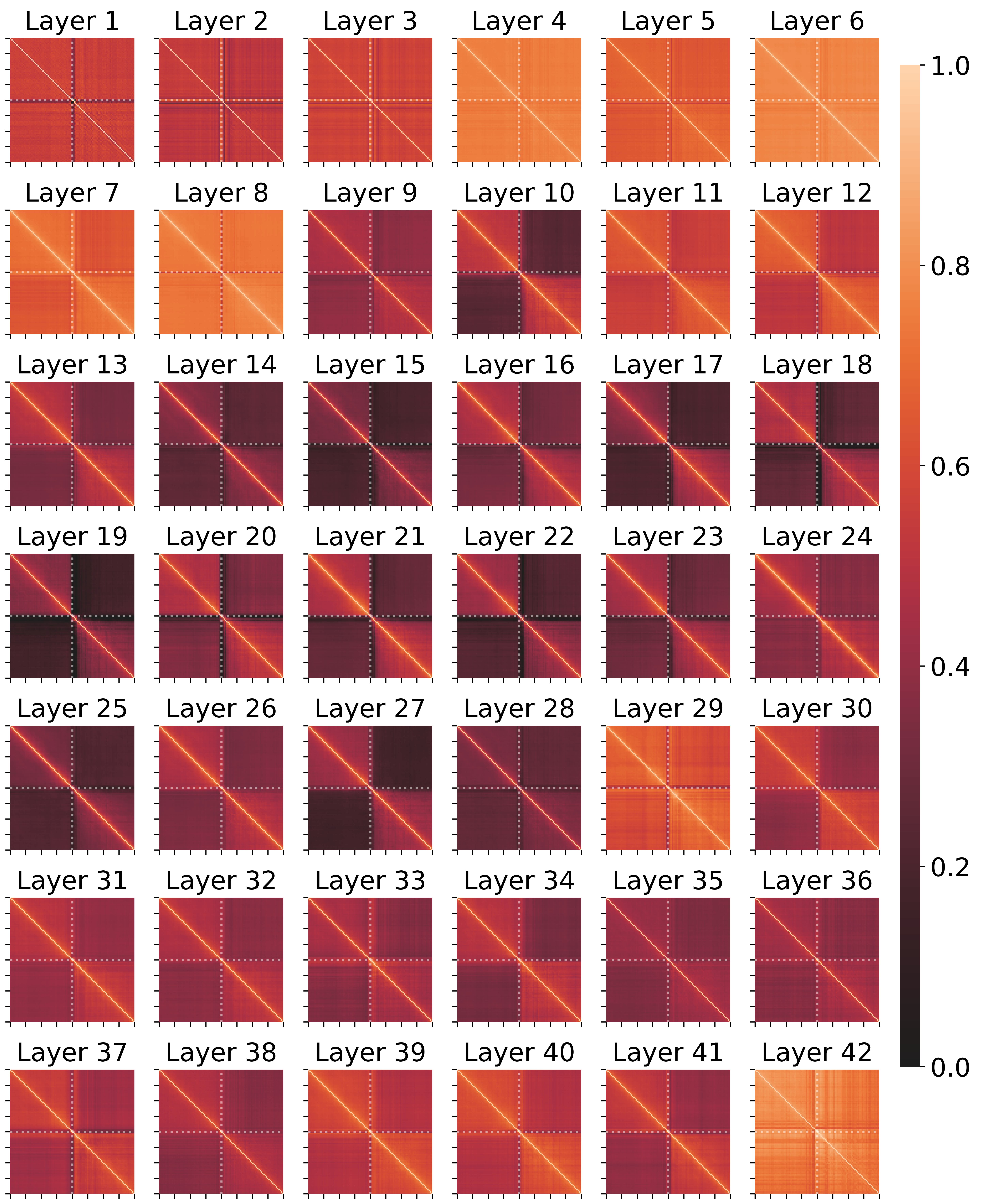}

\caption{Cosine similarity between attention activations across all 42 layers of the model. Results averaged over 1,000 model-generated original contexts, sharing a common structure.}
\label{fig:cosinesim42}
\end{figure}

\begin{figure}[ht]
\centering
\includegraphics[width=0.49\textwidth]{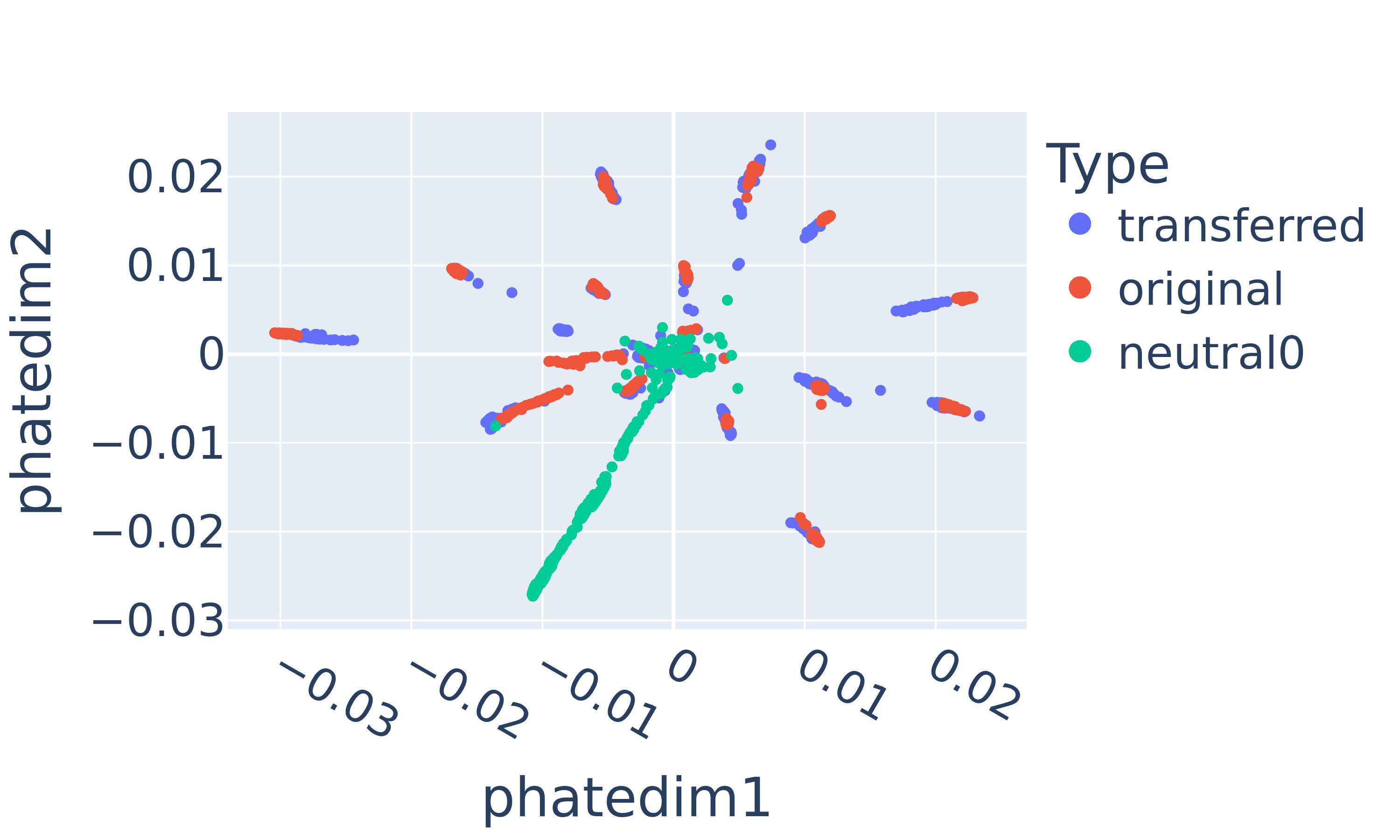}
\includegraphics[width=0.49\textwidth]{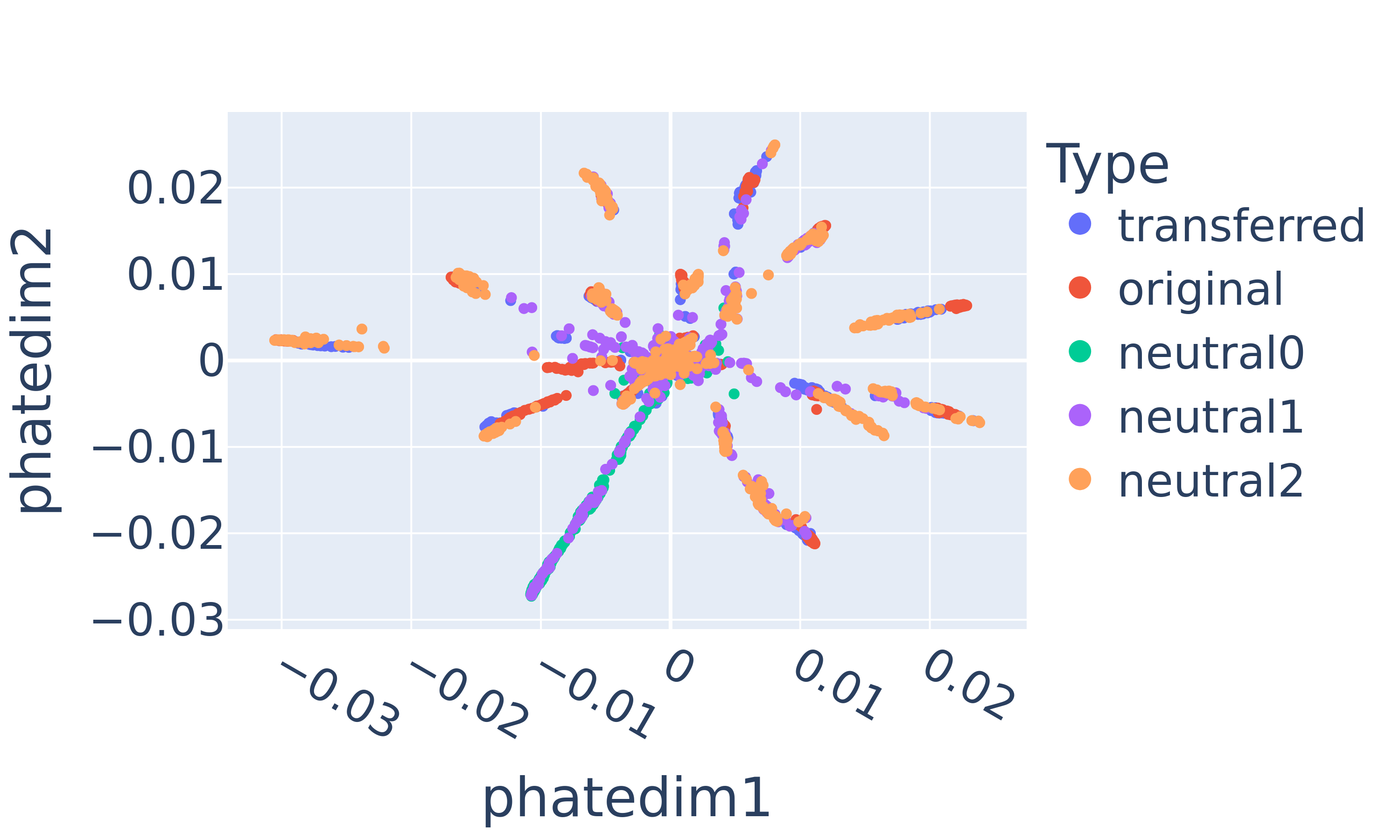}
\caption{Comparison of context similarity between (left) original, neutral0, and activation-transferred text generations and (right) adding neutral1 and neutral2. Contexts were summarized using sentence transformer embeddings and visualized with PHATE, with results shown over 1,000 model-generated original contexts.}
\label{fig:phate}
\end{figure}

\newpage

Below is a table that summarizes the results in Figure \ref{fig:violin}. To compare the neutral2 and transferred, we conducted a T-test on the two sets of cosine distance. This returned a t-statistic of -8.79, with $p = 3.19 \times 10^{-18}$.

\vspace{0.5cm}

% \begin{table}
% \centering
% \resizebox{\textwidth}{!}{%
% \begin{tabular}{lllll}
% \hline
% Type of Generation &
%   \textbf{Neutral0} &
%   \textbf{Neutral1} &
%   \textbf{Neutral2} &
%   \textbf{Transferred} \\ \hline
% \textbf{\begin{tabular}[c]{@{}l@{}}Mean cosine distance \\ to the original generation (Std)\end{tabular}} &
%   0.973 (0.06) &
%   0.616 (0.293) &
%   0.303 (0.239) &
%   0.214 (0.210) \\ \hline
% \end{tabular}%
% }

% \caption{Mean cosine distances to the original generation with standard deviations. Contexts are summarized using sentence transformers, and distributions were taken over 1,000 original contexts.}
% \label{tab:furtherres_avg}
% \end{table}

\begin{table}[h]
\caption{Mean cosine distances to the original generation with standard deviations. Contexts are summarized using sentence transformers, and distributions were taken over 1,000 original contexts.}
\label{tab:furtherres_avg}
\centering
\resizebox{0.85\textwidth}{!}{%
\begin{tabular}{lllll}
\toprule
Type of Generation & \textbf{Neutral0} & \textbf{Neutral1} & \textbf{Neutral2} & \textbf{Transferred} \\
\midrule
\begin{tabular}[t]{@{}l@{}}Mean cosine distance \\ to the original generation (Std)\end{tabular} &
  0.973 (0.06) & 0.616 (0.293) & 0.303 (0.239) & 0.214 (0.210) \\
\bottomrule
\end{tabular}%
}
\end{table}

\textbf{Experimenting with more complex topics}

To test whether our transition token only contains information for the immediate next 1-2 tokens, we try experimenting with more complex topics.  We have not yet run the full experiments for these cases, but anecdotally trying with dozens of examples shows that the double newline token only contains enough information to capture a part of the topic.

\textbf{Topic: Alternative energy storage}

\textbf{Original output:} 'Alternative energy storage is now a crucial topic ' 

\textbf{Start of transferred output 1}
Alternative energy sources are becoming increasingly important in the face of climate change and the growing need for sustainable energy solutions. These sources, such as solar, wind, and geothermal...

\textbf{Start of transferred output 2}
Alternative energy sources are becoming increasingly important as the world transitions towards a more sustainable future.  Solar panels, wind turbines, and other renewable energy technologies are ...

\textbf{Start of transferred output 3}
Alternative energy sources are becoming increasingly popular as the world seeks to reduce its reliance on fossil fuels. Solar panels, wind turbines, and hydroelectric dams are just a few examples ...

\textbf{Topic: Astrophysics recent findings}

\textbf{Original output:} 'Astrophysics has recently made strides in understanding'

\textbf{Start of transferred output 1}
Recent advancements in artificial intelligence (AI) have led to the development of powerful tools that can analyze and interpret data with unprecedented accuracy.

\textbf{Start of transferred output 2}
Recent advancements in artificial intelligence (AI) have led to the development of sophisticated algorithms that can analyze ...

\textbf{Start of transferred output 3}
Recent research has shown that the use of AI in healthcare is rapidly growing. This is due to the ability of AI to analyze...

Below is one one very unusual case where each of the three completions started with  a disclaimer. Two of which contained correct full phrase.

\textbf{Topic: Renewable energy sources
}
\textbf{Original output:} 'Renewable energy sources are sources of energy that...'

\textbf{Start of transferred output 1}
It is important to note that the information provided here is for general knowledge and should not be considered as a substitute for professional advice.
Renewable energy sources are becoming increasingly popular as a way...

\textbf{Start of transferred output 2}
It is important to note that the information provided here is for general knowledge and should not be considered as professional advice.
Renewable energy sources are becoming increasingly popular...

\textbf{Start of transferred output 3}
It is important to note that the use of AI in healthcare is still in its early stages. While AI has the potential to revolutionize healthcare, it is important to be aware of the ethical...

\section{Experimental details}
\label{sec:exp_details}

For our observational experiment in Section \ref{sec:observing}, note that a few generations were excluded from our analysis as the outputs were not in line with the expected structure. (e.g. text wasn't about two different topics, one blended text about two topics). 

To make the attention weights plot in figure \ref{fig:observation_exp}, we collected the attention weights in each head of a layer and summed them up. We then average the result from each layer and each text to get a combined plot.

\begin{table}[h]
\caption{Computing details}
\centering
\resizebox{\textwidth}{!}{%
\begin{tabular}{lll}
\toprule
Experiment & \textbf{Compute specification} & \textbf{Compute time} \\
\midrule
Observational & 1x RTX A4000 16GB & 6 mins \\
(Section \ref{sec:observing}) & & \\
\addlinespace
Transferred generations & 2x RTX A4000 16GB & 24 hours for generating the 5 types of texts: \\
(Section \ref{sec:experimenting}) & & (original, transferred, neutral, neutral1, neutral2). \\
% & & The rest was negligible. \\
\bottomrule
\end{tabular}%
}
\end{table}

We used the Huggingface Transformers library \cite{huggingface-transformers} to implement the extraction of attention weights.

\section{Social Impact}
\label{sec:social_impact}

Understanding the inner workings of large language models (LLMs) through mechanistic interpretability has both positive and negative societal implications. On the positive side, this research can significantly enhance the safety and reliability of LLMs. By deciphering how these models plan ahead and make decisions, we can develop better methods for detecting and mitigating harmful outputs, such as disinformation, biased decision-making, or unintended behaviors. Improved interpretability can also aid in the development of more robust safety measures, ensuring that LLMs align with human values and ethical standards, and help build trust in AI systems deployed in sensitive applications like healthcare, education, and critical infrastructure.

However, this work also presents potential risks. A deeper understanding of LLMs’ mechanisms can be exploited to design targeted adversarial attacks or to manipulate model outputs in malicious ways. For instance, insights gained could be misused to bypass existing safeguards, generate more convincing disinformation, or optimize models for malicious tasks such as creating fake profiles or enhancing surveillance tools. Additionally, as interpretability techniques become more advanced, there is a risk that they could be used to reverse-engineer proprietary or confidential models, leading to intellectual property theft or unauthorized replication of models.

To mitigate these risks, it is important to develop safeguards alongside interpretability advancements, such as gated access to sensitive findings, rigorous monitoring of model usage, and collaboration with policymakers to create frameworks that ensure the ethical application of these insights. Balancing transparency with security will be crucial in leveraging the benefits of mechanistic interpretability while minimizing its potential misuse.

\newpage
\section*{NeurIPS Paper Checklist}

\begin{enumerate}

\item {\bf Claims}
    \item[] Question: Do the main claims made in the abstract and introduction accurately reflect the paper's contributions and scope?
    \item[] Answer: \answerYes{}
    \item[] Justification: Yes, we do demonstrate that patching token activations can transfer significant information about the context of the following paragraph.

\item {\bf Limitations}
    \item[] Question: Does the paper discuss the limitations of the work performed by the authors?
    \item[] Answer: \answerYes{} % Replace by \answerYes{}, \answerNo{}, or \answerNA{}.
    \item[] Justification: A subsection is dedicated to Limitations in the Discussion \ref{sec:limitations}.

\item {\bf Theory Assumptions and Proofs}
    \item[] Question: For each theoretical result, does the paper provide the full set of assumptions and a complete (and correct) proof?
    \item[] Answer: \answerNA{}.
    \item[] Justification: There are no theoretical results in this paper.

    \item {\bf Experimental Result Reproducibility}
    \item[] Question: Does the paper fully disclose all the information needed to reproduce the main experimental results of the paper to the extent that it affects the main claims and/or conclusions of the paper (regardless of whether the code and data are provided or not)?
    \item[] Answer: \answerYes{}
    \item[] Justification: We provide a link to an anonymous repository with our code along with the prompts for all our experiments. The language model we used -Gemma 2, 9b- is cited and open-source.

\item {\bf Open access to data and code}
    \item[] Question: Does the paper provide open access to the data and code, with sufficient instructions to faithfully reproduce the main experimental results, as described in supplemental material?
    \item[] Answer: \answerYes{}
    \item[] Justification: Code and data are provided or open-source.
    \item[] Guidelines:
    \begin{itemize}
        \item The answer NA means that paper does not include experiments requiring code.
        \item Please see the NeurIPS code and data submission guidelines (\url{https://nips.cc/public/guides/CodeSubmissionPolicy}) for more details.
        \item While we encourage the release of code and data, we understand that this might not be possible, so ???No??? is an acceptable answer. Papers cannot be rejected simply for not including code, unless this is central to the contribution (e.g., for a new open-source benchmark).
        \item The instructions should contain the exact command and environment needed to run to reproduce the results. See the NeurIPS code and data submission guidelines (\url{https://nips.cc/public/guides/CodeSubmissionPolicy}) for more details.
        \item The authors should provide instructions on data access and preparation, including how to access the raw data, preprocessed data, intermediate data, and generated data, etc.
        \item The authors should provide scripts to reproduce all experimental results for the new proposed method and baselines. If only a subset of experiments are reproducible, they should state which ones are omitted from the script and why.
        \item At submission time, to preserve anonymity, the authors should release anonymized versions (if applicable).
        \item Providing as much information as possible in supplemental material (appended to the paper) is recommended, but including URLs to data and code is permitted.
    \end{itemize}

\item {\bf Experimental Setting/Details}
    \item[] Question: Does the paper specify all the training and test details (e.g., data splits, hyperparameters, how they were chosen, type of optimizer, etc.) necessary to understand the results?
    \item[] Answer: \answerYes{},
    \item[] Justification: All experimental details can be found in the repository.

\item {\bf Experiment Statistical Significance}
    \item[] Question: Does the paper report error bars suitably and correctly defined or other appropriate information about the statistical significance of the experiments?
    \item[] Answer: \answerYes{}
    \item[] Justification: Yes, in our main experiment \ref{sec:experimenting} we report the standard errors across 1000 generations.

\item {\bf Experiments Compute Resources}
    \item[] Question: For each experiment, does the paper provide sufficient information on the computer resources (type of compute workers, memory, time of execution) needed to reproduce the experiments?
    \item[] Answer: \answerYes{}
    \item[] Justification: We provide these details in Section \ref{sec:exp_details} of the Appendix.
    
\item {\bf Code Of Ethics}
    \item[] Question: Does the research conducted in the paper conform, in every respect, with the NeurIPS Code of Ethics \url{https://neurips.cc/public/EthicsGuidelines}?
    \item[] Answer: \answerYes{}
    \item[] Justification: The NeurIPS Code of Ethics was read and respected.

\item {\bf Broader Impacts}
    \item[] Question: Does the paper discuss both potential positive societal impacts and negative societal impacts of the work performed?
    \item[] Answer: \answerYes{}
    \item[] Justification: Yes, we discuss these in the Appendix, under section \ref{sec:social_impact}.
    
\item {\bf Safeguards}
    \item[] Question: Does the paper describe safeguards that have been put in place for responsible release of data or models that have a high risk for misuse (e.g., pretrained language models, image generators, or scraped datasets)?
    \item[] Answer: \answerNA{}
    \item[] Justification: This is outside of the scope of our paper.

\item {\bf Licenses for existing assets}
    \item[] Question: Are the creators or original owners of assets (e.g., code, data, models), used in the paper, properly credited and are the license and terms of use explicitly mentioned and properly respected?
    \item[] Answer: \answerYes{},
    \item[] Justification: All code and models are explicitly credited.

\item {\bf New Assets}
    \item[] Question: Are new assets introduced in the paper well documented and is the documentation provided alongside the assets?
    \item[] Answer: \answerYes{}
    \item[] Justification: The assets in our repository are well documented.

\item {\bf Crowdsourcing and Research with Human Subjects}
    \item[] Question: For crowdsourcing experiments and research with human subjects, does the paper include the full text of instructions given to participants and screenshots, if applicable, as well as details about compensation (if any)? 
    \item[] Answer: \answerNA{}.
    \item[] Justification: Our paper is not related to human data.

\item {\bf Institutional Review Board (IRB) Approvals or Equivalent for Research with Human Subjects}
    \item[] Question: Does the paper describe potential risks incurred by study participants, whether such risks were disclosed to the subjects, and whether Institutional Review Board (IRB) approvals (or an equivalent approval/review based on the requirements of your country or institution) were obtained?
    \item[] Answer: \answerNA{}.
    \item[] Justification: There are no human study participants in this study.

\end{enumerate}

\end{document}